\documentclass[journal]{IEEEtran}

\ifCLASSINFOpdf
\else
\fi

\usepackage{graphicx}
\usepackage{booktabs}
\usepackage{url}
\usepackage{algorithm}%
\usepackage{algorithmicx}%
\usepackage{algpseudocode}%

\begin{document}

\title{A Plug-in Tiny AI Module for Intelligent and Selective Sensor Data Transmission}

\author{Wenjun Huang*,
        Arghavan Rezvani*,
        Hanning Chen,
        Yang Ni,
        Sanggeon Yun,
        Sungheon Jeong, \\ and Mohsen Imani
\thanks{All authors are with the Department
of Computer Science, University of California, Irvine,
CA, 92697, USA. e-mail: m.imani@uci.edu.}
\thanks{*These two authors contributed equally to this work.}
}

\maketitle

\begin{abstract}
Applications in the Internet of Things (IoT) utilize machine learning to analyze sensor-generated data. 
However, a major challenge lies in the lack of targeted intelligence in current sensing systems, leading to vast data generation and increased computational and communication costs. 
To address this challenge, we propose a novel sensing module to equip sensing frameworks with intelligent data transmission capabilities by integrating a highly efficient machine learning model placed near the sensor. 
This model provides prompt feedback for the sensing system to transmit only valuable data while discarding irrelevant information by regulating the frequency of data transmission. 
The near-sensor model is quantized and optimized for real-time sensor control. 
To enhance the framework’s performance, the training process is customized and a “lazy” sensor deactivation strategy utilizing temporal information is introduced.
The suggested method is orthogonal to other IoT frameworks and can be considered as a plugin for selective data transmission. 
The framework is implemented, encompassing both software and hardware components. 
The experiments demonstrate that the framework utilizing the suggested module achieves over 85\% system efficiency in terms of energy consumption and storage, with negligible impact on performance.
This methodology has the potential to significantly reduce data output from sensors, benefiting a wide range of IoT applications.
\end{abstract}


%
\IEEEpeerreviewmaketitle

\section{Introduction}
\begin{figure*}[!t]%
\centering
\includegraphics[width=1\textwidth]{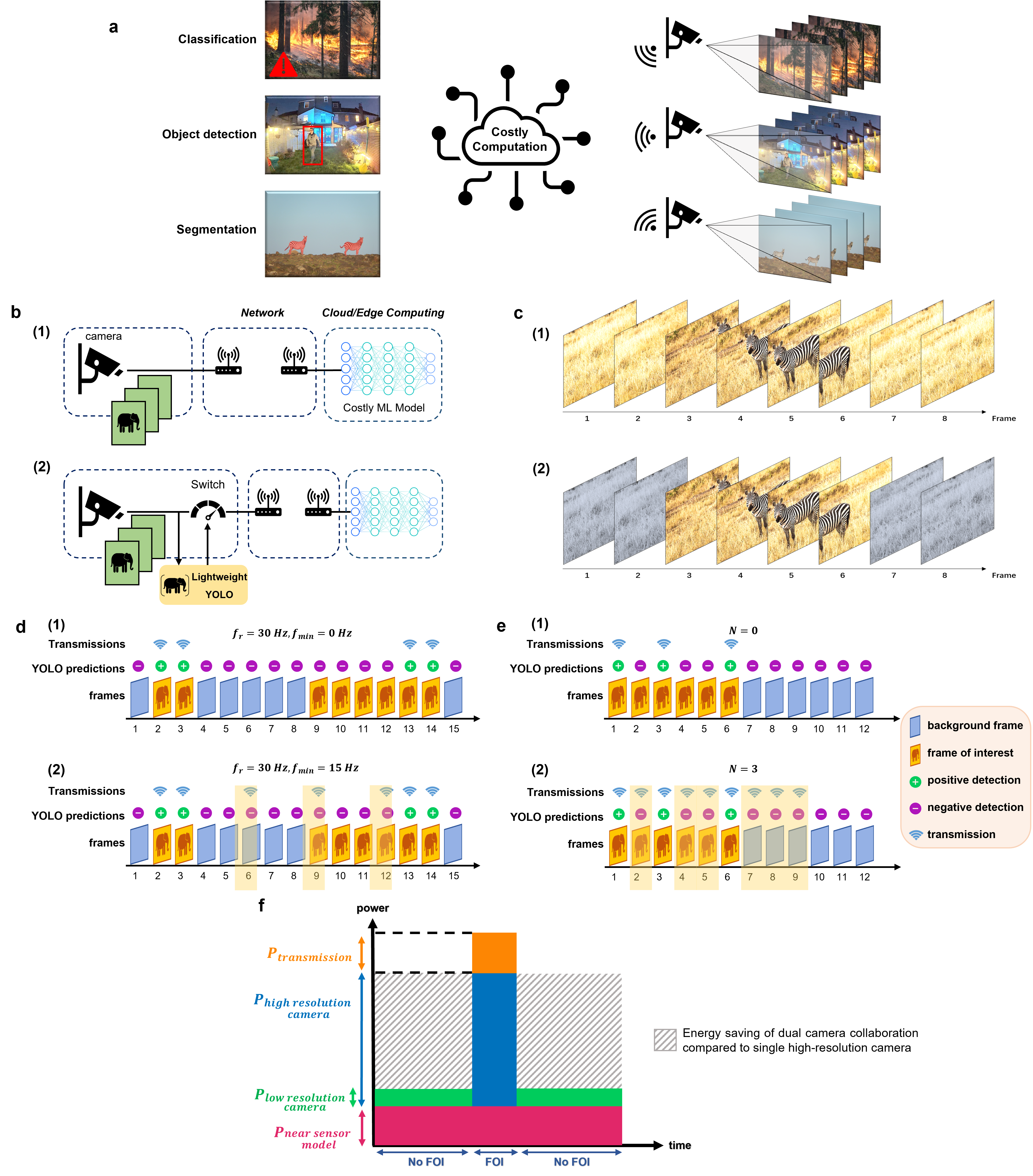}
\caption{\textbf{Motivation and design of our proposed intelligent sensing module.} \textbf{a} Application scenarios of an intelligent system. \textbf{b} General system framework of conventional systems and our system. \textbf{c} Visualization of the data transmission in our system. \textbf{d} Illustration of minimum data transmission frequency (denoted by $f_{min}$) in our system. $f_r$ denotes the camera's refresh rate. \textbf{e} Illustration of lazy sensor deactivation scheme in our system, $N$ is the number for deactivation count. \textbf{f} Energy consumption breakdown on the sensor for the framework utilizing the proposed module using dual-camera collaboration.}\label{figure-1}
\end{figure*}

%
%
%
%
\IEEEPARstart{T}{he} prevalence of ubiquitous sensors is currently experiencing an exponential surge, both in terms of their quantity and the vast amount of data they generate.
Despite the rapid growth, existing approaches to sensor data processing and transmission cannot keep pace due to their algorithmic and architectural limitations \cite{sadeeq2021iot}. 
In numerous IoT applications, data collected by sensors are analyzed using machine learning (ML) models \cite{inbook, 
yang2022iot, djenouri2022sensor, ha2020machine, yun2024hypersense}. 
As the volume of data continues to grow, many applications opt to send the data to more computationally powerful systems, such as edge or cloud computing systems, to execute the learning algorithms. 
In either scenario, a large volume of data is transmitted at a high rate to ensure that all necessary information is captured and processed for various tasks.
The significant amount of data conveyed in both scenarios places high demands on energy and storage resources, resulting in considerable resource pressure and wastage \cite{tsakanikas2023intelligent}. 
This is especially problematic for applications that require a relatively complex and expensive ML model.
Figure~\ref{figure-1}\textbf{a} depicts a typical IoT framework for video monitoring systems, where dense data generated by the camera is continuously analyzed using complex ML models. 
In the framework, visual signals captured by surveillance cameras are transmitted continuously to a costly ML model, which may be hosted on a central server, such as a cloud or edge computing system. 
Depending on the intended purposes, the ML model performs various tasks, including but not limited to classification, object detection, and segmentation \cite{kumari2023deep}.

Many studies attempted to alleviate the energy and storage pressures in IoT applications from multiple perspectives, e.g., computing offloading, resource allocation, etc.
Traditional methods have shown substantial progress in tackling these issues.
Certain research efforts leveraged the Lyapunov optimization algorithm \cite{wang2021event} to identify the optimal decision \cite{chen2018computation, liu2019dynamic}.
Others framed resource allocation and computing offloading as optimization challenges \cite{sun2021cloud, yao2015edal}, employing techniques such as linear programming \cite{chiang2019joint} and mixed-integer nonlinear programming \cite{chen2018task, ning2018cooperative, du2017computation}.
Additionally, traditional approaches like the Alternating Direction Method of Multipliers (ADMM) \cite{wang2019cooperative} and Stackelberg game theory \cite{zheng2018stackelberg} were also investigated. 
However, these approaches exhibit certain limitations. 
Firstly, they require knowledge of the underlying model, which proves challenging due to the intricate and dynamic nature of IoT systems. 
Secondly, they are vulnerable to stuck at local optima, leading to suboptimal efficiency. 
Several more recent research \cite{sun2018deep, yu2017computation,kiran2019joint,ali2019deep,issa2022hyperdimensional, yang2021brainiot} have introduced intelligent offloading strategies grounded in deep learning (DL) decision-making algorithms. These algorithms are employed to determine whether and where tasks should be offloaded for optimal processing.
Furthermore, some research endeavors have placed emphasis on the optimization of hardware structures, thereby enhancing the efficiency of edge computing \cite{biswas2019conv,lammie2019low}.

Different from the work above using ML/DL algorithms to automate offloading and resource allocation, some research proposed solutions to reduce data generated by the sensor.
For example, in the realm of computer vision, 
the approaches aimed at mitigating energy and storage constraints can be categorized into one of two paradigms: \textit{compress-then-analyze} (CTA) and \textit{analyze-then-compress} (ATC) \cite{redondi2013compress}.
Consider video monitoring systems as an example.
In conventional CTA approaches, the acquired visual signals are compressed in pixel domain and spatial domain (e.g., JPEG \cite{wallace1992jpeg}, MPEG-4 \cite{schafer1998mpeg}, or H.264 \cite{wiegand2003overview}) and then transmitted.
However, mild compression can still produce a substantial volume of data in the long run. 
Conversely, aggressive compression methods may result in considerable quality degradation, leading to unsatisfactory results in subsequent analyses \cite{zabala2011effects, tsifouti2012methodology, bagdanov2011adaptive}. 
The ATC approach presents an alternative strategy in which front-end devices extract and transmit features to a central server.
Depending on the specific scenario in which it is being applied, ATC approaches utilize a variety of traditional feature extraction algorithms, ranging from handcrafted methods (e.g., \cite{lowe2004distinctive, bay2006surf, leutenegger2011brisk}) to information theory-based methods \cite{tishby2000information, xiang2011compressed}.
In recent years, more advanced deep-learning-based methods have garnered significant attention.
Several early layers of DNN are deployed on the front-end devices for extracting highly compact and representative features.
In the face recognition task, for example, the face of an individual can be represented by features with several hundred dimensions \cite{schroff2015facenet, sun2015deepid3, ghosh2021edge}.
By representing data in such features, the amount of data that needs to be transmitted can be significantly decreased. 
Additionally, only a few lightweight operations are required to be performed on the central server.

However, one of the weaknesses of such DNN-based ATC methods is their limited generalization ability.
As DNNs are carefully designed, the features extracted and transmitted to the central server are often highly abstract and tailored to that specific task.
However, a given visual signal that contains valuable information, for it to be thoroughly analyzed, often undergoes a series of downstream tasks.
Therefore, the challenge lies in the lack of generalization, making it difficult to design a backbone network capable of extracting suitable features for all these tasks.
In addition, in numerous scenarios, it is necessary to store visual signals for subsequent analysis or future reference. 
The transmission of highly abstract features exceedingly complicates the task of reconstructing the original visual signal on the server side.
Although front-end devices are capable of storing the original signals, the limited storage capacity of these devices presents a predicament.

On the other hand, all the efforts mentioned above, whether from an IoT or ML perspective, still need to process all the data generated from the sensor, neglecting the fact that in many IoT applications (e.g., fire alarm, wildlife monitoring, crime surveillance \cite{yogameena2019deep}, healthcare \cite{huang2021exploration}), only a small fraction of sensor activity typically contains valuable information.
Hence, it is unnecessary to run a costly service, such as a large-scale DNN model, that handles a continuous and complete stream of sensor data, whether on the edge or in the cloud. This is because the service specifically targets only that small fraction of valuable data, yet it still requires processing substantial amounts of irrelevant information.

Observing the limitations of the paradigms previously discussed, in this paper, we rethink and redesign the sensing system framework, proposing a new paradigm that is orthogonal to previous research directions.
The motivation primarily stems from the observation that biological sensors produce data volumes orders of magnitude smaller through the facilitation of intelligent sensing. 
For instance, the human visual system: the retinal system filters and pre-processes a substantial portion of the data, subsequently transmitting key information in the form of spike events to the cortex for processing and learning \cite{dodda2022all, moioli2020neurosciences, mehonic2022brain}.
Drawing inspiration from this observation, we replicate the analysis conducted by the human visual system at the functional level in our system.
Therefore, rather than reducing the data representation or making determinations about where and how data should be relocated for service execution, our approach focuses on reducing the amount of data sent to the system by identifying valuable information.
Our method, acting as a data generation ``filter'', can be applied before any previously mentioned approaches, and easily be integrated into any system as a plug-in.

In our approach, we deploy a lightweight deep learning model near the sensor to detect whether a frame contains useful information, which we refer to as a frame of interest (FOI).
In contrast to the ATC approaches, our approach does not transmit abstract features of frames, but only transmits the original frames containing useful information.
While the size of data transmitted per frame may be larger compared to the ATC approaches, our approach involves the transmission of fewer frames, ultimately resulting in a reduction in the overall amount of data transmitted.
For instance, in the context of object detection, the ML model located near the sensor ensures the sensor only generates data for possible frames/scenes that include objects of interest. In this manner, we can mitigate the huge amount of unnecessary analysis of costly ML models over the central server.
Although this process can also be deployed before the costly ML models at the same place, our method offers substantial savings in transmission costs, encompassing energy, bandwidth, and more.
To enable intelligent sensing, the near-sensor ML model should be fast enough to process frames in real-time and provide feedback for the sensing module \cite{yang2022lead, li2021random, jia2022low}. With the help of this feedback, the sensing module is also aware of the target task of the costly ML model; therefore, it can produce selective and sparse data or even perform some preprocessing on the frames before data transmission. 
Furthermore, we enhance the overall performance of the framework by introducing several effective strategies to mitigate potential misdetections of the lightweight model, which we will explore in the following sections.

In this work, we describe the following contributions:
\begin{itemize}
    \item We propose a new paradigm that improves IoT system energy and storage efficiency orthogonal to the previous paradigms. 
    It can be readily inserted into any existing system, serving as an intelligent data generation ``filter''. We call the sensor exploiting this approach an ``intelligent sensor'' in the rest of the paper.
    \item To illustrate our method, we design a modified DNN model tailored to near-sensor computing for object detection tasks.
    \item We introduce schemes for alleviating possible misdetections of the near-sensor model, including non-zero minimum transmission frequency and lazy deactivation. We also conduct a thorough investigation into their impact on the overall system performance.
    \item We implement the framework encompassing both software and hardware components.
    Our experiments demonstrate that utilizing our intelligent module leads to a substantial reduction in energy and storage consumption in sensing frameworks.
\end{itemize}

\section{Methods}
\subsection{System Framework}
Figure~\ref{figure-1}\textbf{b} illustrates the conventional approach and our proposed sensing and information processing approach. In Figure~\ref{figure-1}\textbf{b}(1), the conventional sensor captures and transmits all the frames to the costly models, regardless of the presence of useful information in the frames.  On the contrary, our approach, as shown in Figure~\ref{figure-1}\textbf{b}(2), utilizes a near-sensor lightweight learning model (in this case a customized YOLO \cite{redmon2016you}) to detect FOIs, and one switch near the sensor controls the data transmission. 
Specifically, the camera captures a continuous stream of frames, which are then fed to the lightweight model for real-time predictions. 
    With the presence of objects of interest inside the frames (the frame is detected as FOI), the switch raises the data transmission frequency, and the frames are transmitted to the central server for more sophisticated operations; if the frame does not contain the objects of interest (detected as a background frame), the switch will turn off and no data transmission occurs. Figure~\ref{figure-1}\textbf{c} provides a visualized example, where the transmitted frames are presented in color while the discarded frames are shaded in gray. 
    Our approach, as demonstrated in Figure~\ref{figure-1}\textbf{c}(2), outperforms conventional approaches depicted in Figure~\ref{figure-1}\textbf{c}(1) by exclusively transmitting frames containing a zebra, resulting in a reduction of storage and energy consumption by half in this particular instance.
    
    Our approach mitigates the energy and storage challenges encountered in IoT applications with the help of the collaboration of a near-sensor lightweight ML model and a data transmission switch. 
    By transmitting only the necessary FOIs to the central server for processing, we reduce the energy consumption of the primary source of energy consumption, i.e., the complex ML model.
    This reduction is achieved while introducing only a negligible energy overhead associated with the near-sensor model. 
    This is in contrast to previous methods that would transmit all frames to the server based on the camera's refresh rate, resulting in significant energy waste due to performing inference on numerous unnecessary frames.
    
    In this work, we concentrate on the effect of the intelligent sensor on energy consumption reduction.
    Each element of the intelligent sensor is elaborated on in the following sections.

    \subsection{Near-sensor Detector}
    \label{subsec: near-sensor detector}
    The near-sensor model is tasked with distinguishing FOIs from all other frames. 
    One way to tackle this problem is by using a classifier. 
    However, the frames captured by a sensor may contain multiple objects of interest with varying scales and positions, while classifiers are typically trained on images that contain a single, centered object (such as those found in CIFAR-10, CIFAR-100 \cite{krizhevsky2009learning}, and ImageNet \cite{deng2009large}). 
    These classifiers have limitations in detecting multiple objects with varying scales and positions. As a result, a deep object detection model is often employed instead.
    Among different object detection models, YOLO, a single-stage detector, is selected.
    Compared with two-shot detectors (e.g., R-CNN, Fast R-CNN, Faster R-CNN, R-FCN \cite{girshick2014rich, girshick2015fast, ren2015faster, dai2016r}), YOLO is lightweight, faster, and with comparable accuracy in a suitable scenario \cite{deng2023lightweight, zahrawi2023improving}.
    These features make YOLO a good candidate for being embedded into the sensor.

    The output layer of YOLO contains bounding box predictions concatenated to the class prediction and objectness confidence. 
    However, the goal of our intelligent sensor is to detect the existence of objects of interest, regardless of their position in the frame.  
    Therefore, we can only keep the objectness confidence in the YOLO output, which can be used further to determine FOI.
    We set a threshold for the objectness confidence, and only the frames with a confidence value exceeding this threshold are transmitted.
    As we increase the threshold, the detection becomes stricter, resulting in fewer frames being considered FOI and more frames being classified as background.

    Our framework's definition allows us to customize the YOLO model in the following ways:
        \subsubsection{Model Optimization}
        The architectures of YOLO series contain several repetitive blocks. 
        Although these blocks contribute to the model capacity, they make the model power-hungry and slower during inference.
        For example, YOLOv5 model family has five variations: x-large, large, medium, small, and nano. 
        While each model shares the same structure, they differ in the network's depth and the number of filters in different layers (width).
        Since our model does not predict bounding boxes, we can modify its depth and width to create a more lightweight model that still achieves comparable performance on our task.

        As mentioned earlier, the output of YOLO model not only contains the confidence but also concatenates the bounding box information, which is not required in our proposed module. 
        Consequently, we can remove the part of the model associated with the bounding box during the inference to reduce the model size. 
        \subsubsection{Inference Simplification}
        YOLO utilizes non-max suppression (NMS) algorithm as the final step to pick the most appropriate bounding box for the object among all of the predicted boxes for that specific object. 
        NMS algorithm starts with selecting the box with the highest objectness score among all, removing all the boxes with high overlap with the selected box, and repeating these steps iteratively. 
        However, since the sensing scenario does not require bounding boxes, we can simplify this step.
        Instead of running the NMS algorithm, we only keep the highest objectness confidence.
        If there is one confidence value greater than the threshold, it indicates the presence of at least one object in the prediction.
        Therefore, by solely comparing the highest confidence value with the threshold, we can achieve comparable performance while reducing the near-sensor inference time.
        \subsubsection{Model Quantization \& Loss Function Customization}
        Model quantization is another well-known approach to accelerating model inference. 
        It involves using fewer bits to store model parameters while maintaining nearly the same level of accuracy
        \cite{alqahtani2021literature, goel2020survey, coelho2021automatic,chakraborty2020constructing}.
        Aggressive quantization leads to a highly lightweight model, but at the cost of reduced accuracy compared to the original model. 
        On the other hand, less aggressive quantized models experience minimal accuracy loss, but they are not as lightweight as aggressively quantized models
        \cite{wu2020integer}.
        The amount of tolerable accuracy loss varies across different tasks. 
        
        Moreover, refining the loss function can enhance the performance of the model when subjected to intensive quantization.
        The conventional YOLOv5 has three loss terms: 
        \begin{equation}
            L = l_{obj} + l_{cls} + l_{bbox}
            \label{eq: yolo loss}
        \end{equation}
        where $l_{obj}$, $l_{cls}$, and $l_{bbox}$ are objectness confidence loss, classification loss, and bounding box loss, respectively.
        Among the loss terms, reducing the $l_{obj}$ and $l_{cls}$ loss terms contributes to accurate object detection and classification, resulting in improved performance of our model. 
        In contrast, the $l_{bbox}$ loss term, which corresponds to the precise bounding box position, has a negative impact on our model. 
        This is because it forces the model to make a compromise during the gradient descent search, making it more difficult for the model to converge to the optimal. 
        By removing the $l_{bbox}$ term, our near-sensor model can prioritize the detection of FOIs without considering the bounding box generation, enabling the model to achieve a higher degree of quantization while maintaining a comparable level of accuracy.

\subsection{Data Transmission Frequency}
    The intelligent sensor's embedded switch regulates the frequency of data transmission, thereby reducing the volume of data transmitted to the central server. 
    If the camera records FOIs, the switch should be configured to transmit all FOIs to the server, with a frequency equivalent to the camera's refresh rate. 
    Conversely, when the camera captures background frames, the switch should lower the data transmission rate to save energy. 
    This reduced frequency is referred to as the \textit{minimum transmission frequency}.
    The minimum transmission frequency can vary between zero and the camera's refresh rate.
    If the minimum transmission frequency is set to match the camera's refresh rate, all frames captured by the camera are forwarded to the server, indicating that the switch is unaffected by the predictions of the lightweight model. 
    In this scenario, the volume of data transmitted to the server is identical to that of conventional systems. 
    Conversely, when the minimum transmission frequency is set to zero, any frames identified as background frames would not be transmitted to the server. Figure~\ref{figure-1}\textbf{d}(1) demonstrates the data transmission of our proposed system.
    In the figure, the blue frames represent the background and the yellow frames with an elephant depict FOIs. 
    A positive or negative sign is used to present the near-sensor model predictions;
    The frames that are marked with a positive sign represent the prediction of FOIs. 
    The frames being transmitted to the server are indicated by the Wi-Fi icon.
    Only the frames that are recognized as FOI (frames 2, 3, 13, and 14), are transmitted.

    However, even though the lightweight model displays a high level of accuracy, it is still inevitable to misdetect some FOIs as background frames,
    and these misdetected frames are all discarded when the minimum transmission frequency is zero since the data transmission is completely halted.
    This wrong discard can be alleviated by increasing the minimum transmission frequency, which means that even if the frames are detected as background frames, they are still transmitted to the server regularly at a lower non-zero frequency.
    An example demonstrating the effect of increasing the minimum transmission frequency is shown in Figure~\ref{figure-1}\textbf{d}(2). 
    When the minimum transmission frequency equals zero (Figure~\ref{figure-1}\textbf{d}(1)), all the frames detected as background are discarded by the intelligent sensor. 
    This significantly reduces the amount of data transmitted while also losing some useful information (e.g., frames 9 - 12).
    To reduce the number of missing FOIs, we increased the transmission frequency in Figure~\ref{figure-1}\textbf{d}(2).
    In the figure, the camera's refresh rate $f_r = 30$ Hz, and the minimum transmission frequency $f_{min}$ is set to $ f_r / 2$ (i.e., $15$ Hz). Under this setting, even if the transmission frequency is tuned down, the sensor would also send one frame every two frames.
    From the figure, we can observe that although the prediction of the lightweight model maintains the same, we transmit more FOIs to the server (frame 9 and frame 12).

    \subsection{Lazy Sensor Deactivation}
        Since FOIs contain valuable information, in this work, the priority is given to transmitting all FOIs rather than mistakenly transmitting a background frame.
        Therefore, we define misdetections as the FOIs which are not transmitted. 
        Considering the fact that the frames in a video have temporal correlation, we assume that if the camera captures an FOI, the following frame is likely to be an FOI as well.
        Thus, in order to reduce the misdetection of FOIs, inspired by \cite{caravagna2012lazy}, we proposed a scheme for lazy sensor deactivation, which considers the detection results of neighboring frames.
        However, unlike the work in \cite{caravagna2012lazy} which schedules observation points over the target execution, our scheme entails monitoring the number of consecutive background frames detected by the near-sensor model.
        The switch maintains a high transmission frequency until the count ($C_1$) of consecutive background detection reaches a pre-defined number ($N$).
        Once the number is met, the switch tunes down the transmission frequency and resets the count.
        The count is reset to zero whenever an FOI is identified.
        The adoption of our lazy sensor deactivation scheme enables the detector to rectify the misdetection of a single frame by utilizing the adjacent frame's information.
        In comparison to the detector without the lazy sensor deactivation scheme, utilizing our approach preserves more FOIs since an occasional misdetection cannot affect the transmission frequency. 
        Decisions for tuning the transmission frequency are made based on a few adjacent frames.
        Figure~\ref{figure-1}\textbf{e} provides an example that demonstrates the advantages of implementing lazy sensor deactivation. 
        In this example, the value of $N$ is set to 3. 
        When compared to the system that does not utilize lazy deactivation (shown in Figure~\ref{figure-1}\textbf{e}(1)), the implementation of lazy deactivation (Figure~\ref{figure-1}\textbf{e}(2)) also transmits the FOIs that are misdetected by the near-sensor model, as demonstrated by the transmission of frames 2, 4, and 5.
        
        The utilization of the lazy sensor deactivation scheme incurs two costs from a storage perspective. 
        The first cost arises when a negative sample is mistakenly identified as an FOI, leading to the reset and restart of the count. 
        In the worst-case scenario, a single negative sample misdetection results in storing $2N$ additional frames. 
        Nonetheless, we find this cost acceptable as our primary concern lies in preserving the completeness of FOIs. 
        The inclusion of a few extra negative samples following FOIs does not influence the pertinent information we aim to retain. 
        Moreover, given that $N$ is not an excessively large value, our storage capacity can handle these rare occurrences of misdetection.

        The second cost inherent in the lazy sensor deactivation scheme manifests in the recovery of the transmission frequency to a high level and the subsequent repetition of counting following each period of frequency decrease. 
        Given that a large proportion of frames comprise background and such frames often appear in the form of segments, the frames following a low transmission frequency period are more likely to be background as well. 
        As a result, in a long sequence of background frames, the detector stores $N$ more frames after each low transmission frequency period. 
        To mitigate the redundancy following each period, we introduced one more count ($C_2$) to monitor the number of consecutive low transmission frequency periods. 
        This count is used to calculate $N_{new}$ for consecutive background frames:
        \begin{equation}
           N_{new} =\max(1, \frac{N}{2^{C_2}}) 
            \label{eq: threshold}
        \end{equation}
        Upon tuning down the transmission frequency, $C_2$ increments by 1. 
        However, the detection of an FOI interrupts the consecutive low transmission frequency periods, resetting the count $C_2$ to zero. 
        At the start of each period, equation~(\ref{eq: threshold}) determines the number ($N_{new}$) for that particular period. 
        Using the count $C_2$ for consecutive low transmission frequency periods gradually decreases the threshold from $N$ to $1$ in the long run, leading to greater storage and energy savings than the vanilla scheme.
        
        Algorithm~\ref{algoA1} outlines the pseudocode of our approach, which incorporates minimum transmission frequency and lazy sensor deactivation.
        The pseudocode~\ref{code:line6} - \ref{code:line12} indicate the code for lazy sensor deactivation and \ref{code:line14} - \ref{code:line19} indicate the code for minimum transmission frequency, where $f_r$ is the camera's refresh rate, $f_{min}$ is the minimum transmission frequency, and $C_3$ is a count used to determine whether a frame should be transmitted in a minimum transmission frequency period.
    \subsection{Dual-camera Collaboration}
    
    Recording and analyzing valuable information necessitate the utilization of images with high resolution, thereby engendering a predilection for high-resolution cameras within the relevant context. 
    Nevertheless, the sustained operation of such cameras for near-sensor computing proves to be energy burdensome, given their elevated power consumption. 
    Our objective is to furnish dependable performance while concurrently minimizing energy consumption. 
    In pursuit of this goal, we integrated an additional low-resolution, and power-efficient, camera into the sensor configuration for dual-camera collaboration.

    During periods devoid of FOIs, the high-resolution camera remains either inactive or in an idle state to conserve power, while the low-resolution camera is engaged in executing the near-sensor model, as elucidated in Section \ref{subsec: near-sensor detector}. 
    When an FOI is detected by the near-sensor model utilizing the low-resolution camera, the high-resolution camera is activated, capturing and subsequently transmitting the pertinent frames. 
    During this phase, the low-resolution camera is deactivated, given the superior quality of the frames obtained by the high-resolution camera. 
    
    The power consumption breakdown of using dual-camera collaboration on the sensor side is depicted in Figure~\ref{figure-1}\textbf{f}. 
    The incorporation of a power-efficient low-resolution camera results in significant power savings (highlighted in shadow), even compared to using our module with a single high-resolution camera.
    Given the infrequent occurrence of FOIs, our dual-camera collaboration scheme proves to be highly effective in mitigating energy consumption at the sensor side over an extended duration.

    \subsection{Hardware Implementation}
        The hardware acceleration of our model's inference procedure was implemented on AMD-Xilinx Zynq UltraScale+ MPSoC ZCU104 (ZCU104) using field-programmable gate arrays (FPGAs)~\cite{zynq2023}. FPGAs are semiconductor devices that are based on a matrix of configurable logic blocks (CLBs) connected via programmable interconnects. Through hardware programming (such as Verilog or HLS), we can implement an ML accelerator on FPGA. The host program, executed on the ARM Cortex-A53 processor on the ZCU104's processing system (PS), was developed in Python. The communication between the processing system (PS) and the programmable logic (PL) is established through the AMBA Advanced eXtensible Interface (AXI). Here PS side is a host ARM processor and PL side is a reconfigurable logic. Our architecture design is implemented on the top of PL (reconfigurable logic). 
        
        To leverage hardware acceleration, we utilized the AMD-Xilinx deep learning unit (DPU) intellectual property (IP) as our hardware accelerator on the ZCU104's programmable logic (PL) side. Our model was integrated into the DPU using the Vitis AI framework~\cite{kathail2020xilinx}. Vitis AI is an ML compiler framework developed by AMD-Xilinx that automatically maps ML operations (such as convolution and fully connected layers) into Xilinx hardware IP. The Vitis AI version that we choose is 2.0. Furthermore, the cameras are connected to the host ARM CPU, which facilitates communication with the cloud server.
        TCP protocol is used as the communication protocol.
        The overview of our hardware platform is shown in Figure~\ref{fig:hardware}.

\section{Results}
\subsection{Experimental Setup}
In this work, we trained and evaluated our system in the context of animal detection using the Microsoft Common Objects in Context (MS COCO) dataset \cite{lin2014microsoft}, which is widely used for object detection tasks.
In this context, the images in the dataset were selected and relabeled.
The images containing at least one object belonging to the animal category are considered FOI and are labeled 1. 
The remaining frames are considered background and labeled as 0.
The near-sensor lightweight model detects and transmits FOIs while filtering out the background frames.
The detected frames are transmitted to a more sophisticated model, in our case a well-trained Fast R-CNN  model, to perform advanced operations.
The system is implemented using PyTorch \cite{paszke2019pytorch}.
In accordance with the scenario, we ordered the data in the testset with a specific logic: FOIs and background frames are presented in a fragmented manner, appearing consecutively and alternating with each other.
The frames in fragments are ordered randomly.
\subsection{Model Size}
    \begin{figure*}[ht]%
    \centering
    \includegraphics[width=1\textwidth]{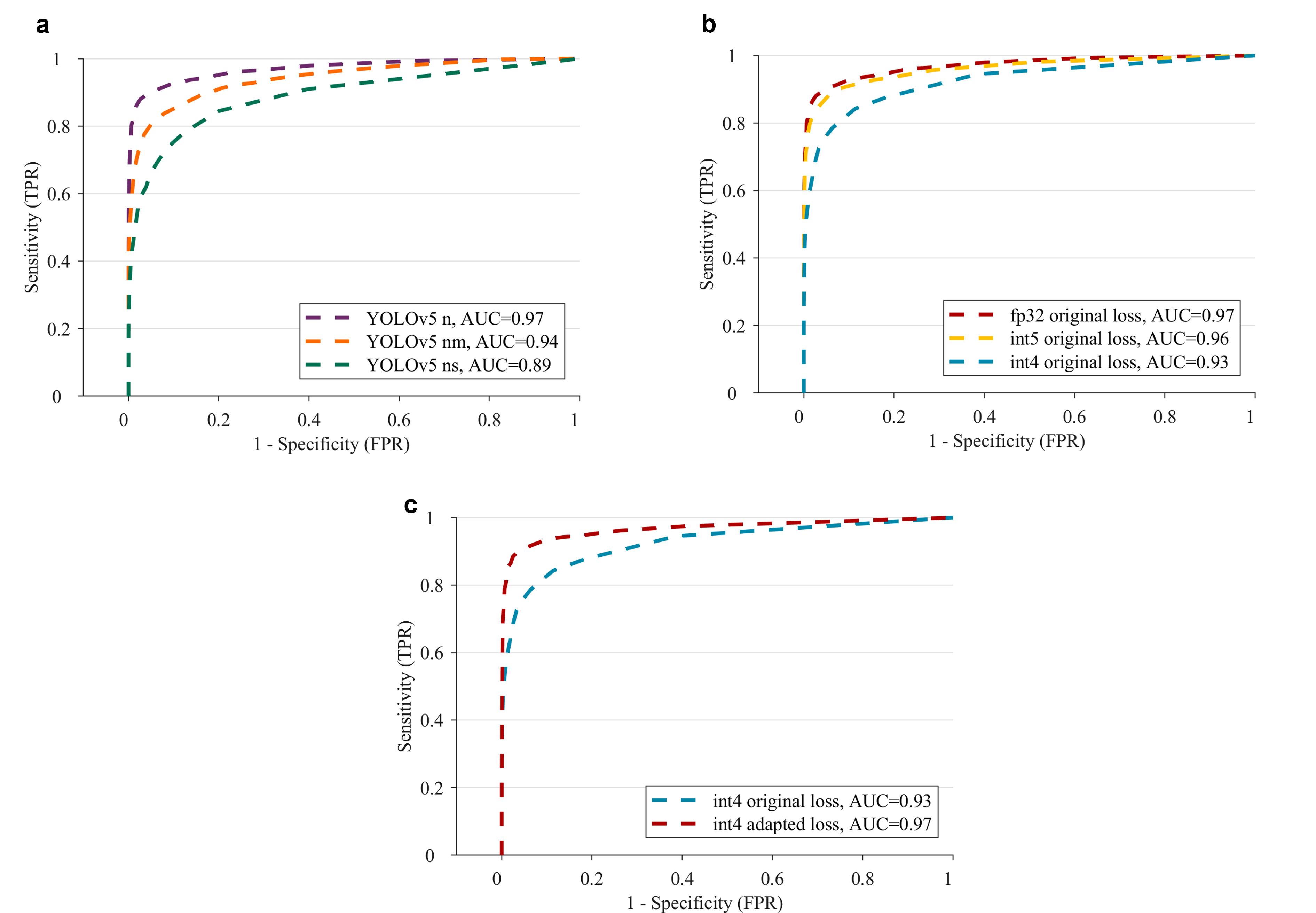}
    \caption{\textbf{In-sensor model comparison.} \textbf{a} ROC curves of three lightweight models. \textbf{b} ROC curves of the models with different quantization trained by original loss. \textbf{c} The ROC curves of the model subjected to int4 quantization, trained with our adapted loss function and the original loss function.}\label{fig: model}
    \end{figure*}
    To begin with, we examined the performance of various lightweight near-sensor models. 
    Specifically, we compared three YOLO-based models, namely YOLOv5n, YOLOv5nm, and YOLOv5ns, sorted in descending order of the number of parameters. YOLOv5n stands for Yolov5 nano, which is the smallest introduced YOLOv5 model. As mentioned earlier in the methods section, we also changed the depth and width (number of filters in different layers) of the YOLOv5n and achieved more lightweight models, which we called nano-medium (YOLOv5nm) and nano-small (YOLOv5ns). In YOLOv5nm, the depth and width are half of the depth and width of the YOLOv5n, and in YOLOv5ns, this ratio is one-third.
    
    We evaluated the trade-off between the sensitivity and specificity of these models using receiver operating characteristic (ROC) curves and area under the curve (AUC), as illustrated in Figure~\ref{fig: model}\textbf{a}. 
    Additionally, Table \ref{table:model_parameters} displays the number of parameters and GFLOPS of each model.
    While the AUC of YOLOv5ns is only slightly lower than that of YOLOv5n, the reduction in model size is significant, with the number of parameters decreasing to only 6.2\% of the latter.
    Furthermore, the detrimental effect resulting from the reduction in model size can be alleviated by incorporating the lazy sensor deactivation scheme and elevating the minimum transmission frequency.

    As mentioned earlier, another approach to obtain more lightweight models is through model quantization. To this end, we utilized the \textbf{kmeans-lut} quantization which is a Look-up-table (LUT) based quantization \cite{cardinaux2020iteratively}, where LUT is generated by K-Means clustering. 
    
    We quantized YOLOv5n into different bit precisions, i.e. 16-bit float point (\textit{fp16}), 8-bit integer (\textit{int8}), 5-bit integer (\textit{int5}), 4-bit integer (\textit{int4}). 
    The performance of both the \textit{fp16} and \textit{int8} quantized models remains unaffected.
    However, as illustrated by Figure~\ref{fig: model}\textbf{b}, when we further reduce bit precision to \textit{int5}, a slight degradation in AUC is observed, and a degradation in performance is noticeable when the model is quantized to \textit{int4}.
    We also assessed the impact of our tailored loss function, which facilitates intensive quantization of the model while maintaining comparable performance levels, as shown in Figure \ref{fig: model}\textbf{c}.

    \begin{table}[ht]
    \centering
    \caption{Model Parameters}
    \begin{tabular}{|c|c|c|}
    \hline
    \textbf{Model Name} & \textbf{No. of Parameters} & \textbf{GFLOPS} \\ \hline
    YOLOv5 n & 1765270 (100\%) & 4.2 \\ \hline
    YOLOv5 nm & 433190 (24.5\%) & 1.1 \\ \hline
    YOLOv5 ns & 108806 (6.2\%) & 0.4 \\ \hline
    \end{tabular}
    \label{table:model_parameters}
    \end{table}

    \subsection{Evaluation Metrics}
    \begin{figure*}[htbp]%
    \centering
    \includegraphics[width=0.86\textwidth]{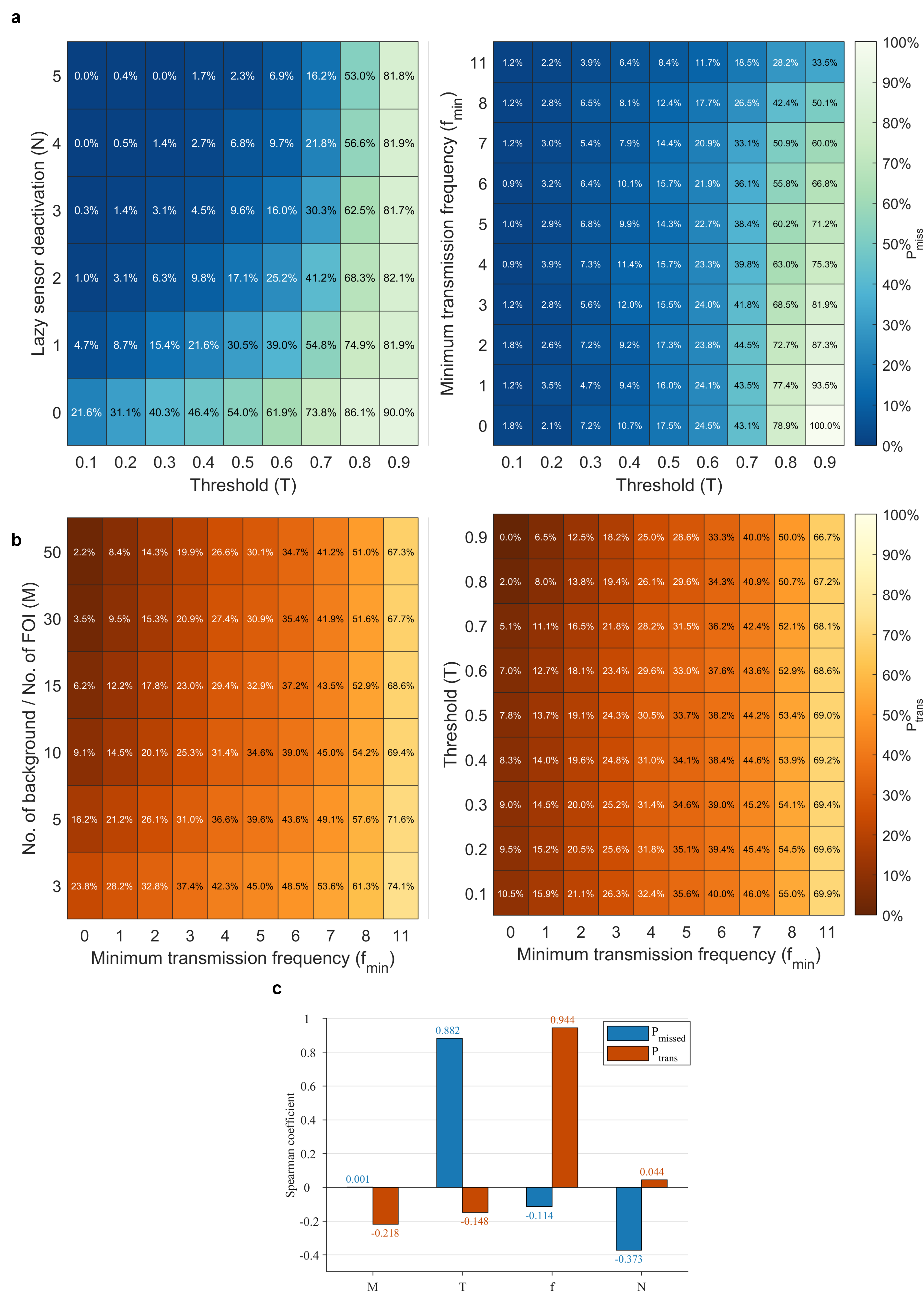}
    
   \caption{\textbf{Performance evaluation.} \textbf{a} Heatmaps that display the miss rate $P_{miss}$ with different parameter combinations (threshold ($T$), the ratio ($M$) of the number of background frames to the number of FOIs, minimum transmission frequency ($f_{min}$), and the count ($N$) at which the sensor deactivates). \textbf{b} Heatmaps that display the percentage of transmission $P_{trans}$ with different parameter combinations. \textbf{c} Spearman coefficient of the parameters.
   }
    \label{fig: results}
    \end{figure*}
    After the discussion of model size, we considered the final goal of the whole intelligent sensing module, which is to detect the animals in all FOIs and implement more sophisticated inference using Fast R-CNN model on the server while minimizing the system energy consumption and occupying minimal storage.
    In essence, our module aims to minimize the miss detection rate, defined as the fraction of FOIs that are not detected by our near-sensor model:
    \begin{equation}
        P_{miss} = \frac{n_{miss}}{n_{FOI}}, 
        \label{eq: missed detection rate.}
    \end{equation}
    where $n_{miss}$ is the number of missed FOIs by the near-sensor model, and $n_{FOI}$ is the total number of FOIs in the video stream.
    
   In addition, we also prioritized the percentage of transmission reduction achieved by our module in comparison to sending all frames captured by the camera, which is defined as:
   \begin{equation}
       P_{trans} = \frac{n_{trans}}{n_{frames}},
       \label{eq: transmission reduction percentage}
   \end{equation}
   where $n_{trans}$ is the number of transmitted frames, and $n_{frames}$ is the total number of frames in the testset. 
   The energy consumption of the framework, including transmission and inference energy of the Fast R-CNN model, is closely related to the number of transmitted FOIs; thus the percentage of transmission reduction serves as a key indicator of the effectiveness of our system. In addition to impacting energy consumption, $P_{trans}$  also reflects the amount of storage that can be saved on the server. 
   
    It is worth emphasizing that our intelligent module involves trade-offs among its various parameters. 
    Altering the values of these parameters can lead to different performances with respect to missed detection frames and the percentage of transmission reduction. 
    For instance, the most extreme scenario is to maintain the transmission frequency equal to the camera's refresh rate and transmit all frames to the server. 
    Although this approach would result in zero missed detection frames, it would also lead to the highest possible energy consumption.
    In the following, we analyze the influence of each parameter on the performance of the framework utilizing the proposed module and discuss the trade-offs between these parameters.

    This study explores the impact of four key parameters on our system's performance: 
        \begin{itemize}
            \item the confidence threshold ($T$) of YOLO
            \item the ratio ($M$) of the number of background frames to the number of FOIs
            \item minimum transmission frequency ($f_{min}$)
            \item the count ($N$) at which the sensor deactivates
        \end{itemize}
    
        Figure~\ref{fig: results}\textbf{a} and \textbf{b} present heatmaps illustrating the impact of the key parameters on $P_{miss}$ and $P_{trans}$.
        For both metrics, a lower value indicates better performance.
        Regarding $P_{miss}$, a higher value of $T$ leads to a notable increase in the number of missed FOIs. 
        However, incorporating a lazier deactivation scheme and increasing the minimum transmission frequency can mitigate the adverse effects associated with a higher value of $T$.
        Conversely, raising the minimum transmission frequency has a predominantly negative impact on $P_{trans}$. 
        However, adopting a high value of $T$ can reduce the amount of data transmission.
        We also examined the relationship between $M$ and $P_{trans}$ and found that our approach exhibits a clear advantage as $M$ increases.
        Specifically, when $M = 50$ (i.e., the number of FOIs accounts for 5\% of the total), our approach can save up to 98\% of storage and transmission energy compared to conventional approaches.
        The Spearman correlation coefficients \cite{spearman1961proof} of the parameters are presented in Figure~\ref{fig: results}\textbf{c}.
        In the figure, a positive coefficient indicates a positive correlation between two variables, while a negative coefficient indicates a negative correlation.
        The magnitude of the coefficient reflects the strength of the association between the variables.
        A trade-off between the parameters $T$, $f$, and $N$ should be considered when aiming to minimize $P_{miss}$.
        Moreover, our approach exhibits superior performance compared to existing methods in terms of $P_{trans}$, particularly when the minimum transmission frequency $f_{min}$ is kept low and FOIs are rare.

\subsection{Energy consumption}

    \begin{table}[]
    \centering
    \caption{Design Acceleration On AMD-Xilinx ZCU104}
    \label{tab:fpga_resource}
    \resizebox{1\columnwidth}{!}{
    \begin{tabular}{l|ccccc}
    \toprule
                               & LUT                        & FF      & \multicolumn{1}{l}{BRAM} & URAM    & DSP                         \\
    \midrule
    \multicolumn{1}{c|}{Total} & 84.9K                      & 146.5K  & 224                      & 40      & 844                         \\
    Available                  & \multicolumn{1}{l}{230.4K} & 460.8K  & 312                      & 96      & 1728                        \\ 
    Utilization                & 36.87\%                    & 31.80\% & 71.79\%                  & 41.67\% & \multicolumn{1}{l}{48.84\%} \\                                                                       
    \bottomrule
    \end{tabular}%
    }
    \end{table}
    
\begin{figure*}[!ht]%
    \centering
    \includegraphics[width=0.8\textwidth]{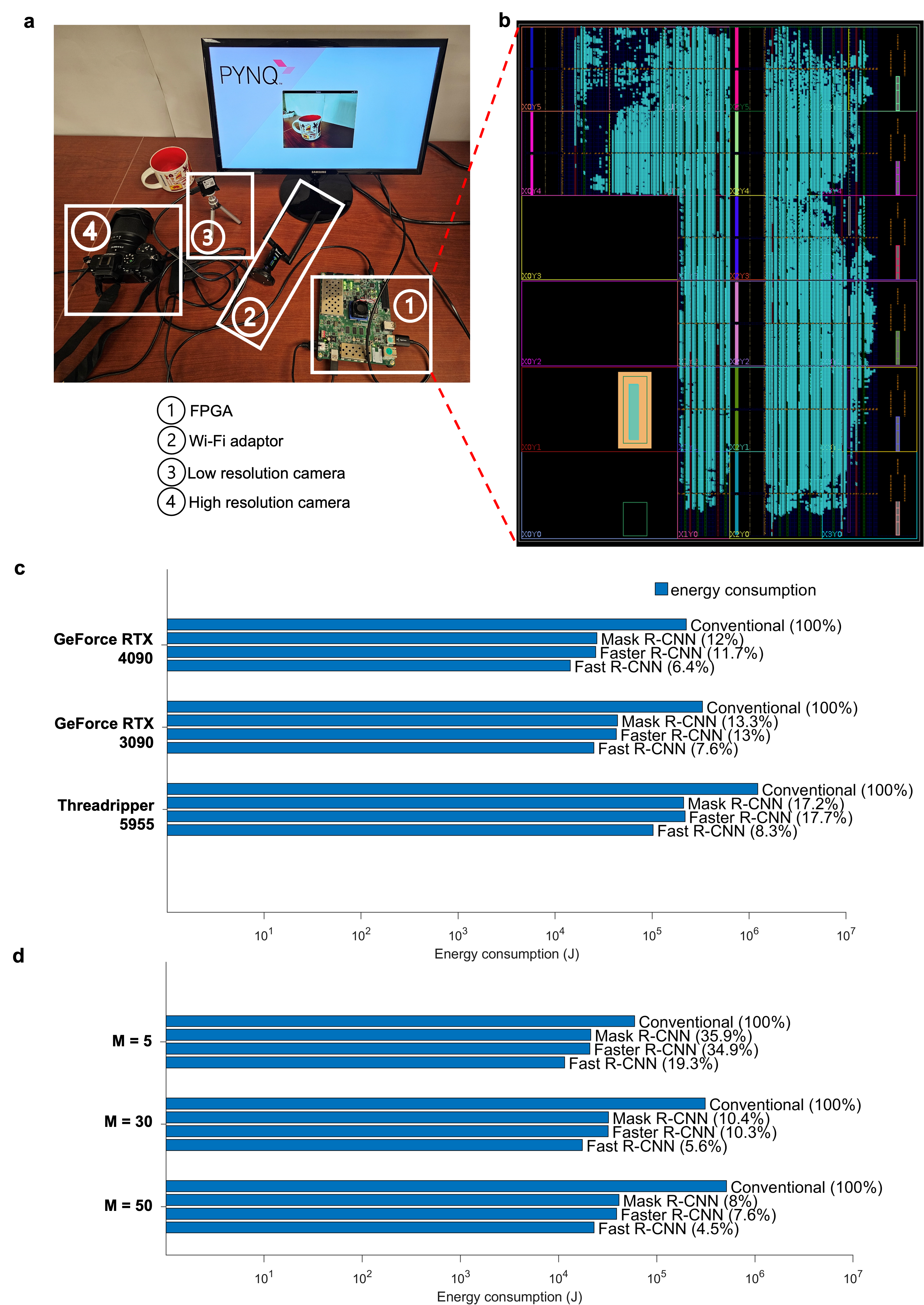}
   \caption{\textbf{Experiment results.} \textbf{a} System built for experiments. \textbf{b} Accelerator placement layout on AMD Xilinx ZCU104 FPGA. \textbf{c} Energy consumption comparison with $M = 20$ (the total number of the frames $n_{total}=21336$, $P_{miss}=3\%\pm0.6\%$). The conventional system implements Fast R-CNN on the server. \textbf{d} Energy consumption comparison with various $M$ values. All servers are equipped with GeForce RTX 4090.}
    \label{fig: system}
    \end{figure*}

    To meet the requirements of the proposed scenario, the near-sensor model is deployed on a resource-limited low-power edge-level FPGA \cite{tang2022ef}.
    The system setup on the sensor side is depicted in Figure~\ref{fig: system}\textbf{a}.
    A high-resolution camera (\textcircled{4}), a low-resolution camera (\textcircled{3}), and a Wi-Fi adaptor (\textcircled{2}) are connected to the FPGA board (\textcircled{1}) via cable to capture and transmit the frames to the server.
    In addition, a screen is utilized for visualizing the information captured by the camera. In Figure~\ref{fig: system}\textbf{b}, we present the accelerator placment layout on AMD Xilinx ZCU104 FPGA. In Table~\ref{tab:fpga_resource}, we present the FPGA resource utilization result.
    
    The energy consumption of conventional sensing systems is composed of three parts: energy consumption on the sensor, transmission energy, and inference energy consumed at the server. 
    In contrast, the framework utilizing our module also includes another part of energy consumption which is near-sensor model energy consumption.
    Despite introducing an additional energy consumption, our approach remarkably reduces the number of frames sent to the server for inference, resulting in significant energy savings at the server, as illustrated in Figure~\ref{fig: system}\textbf{c} and \textbf{d}. 
    Figure~\ref{fig: system}\textbf{c} shows the implementation of three models (i.e., Mask R-CNN \cite{he2017mask}, Faster R-CNN, Fast R-CNN) on three platforms, including two GPUs (GeForce RTX 4090 and GeForce RTX 3090) and one CPU (AMD Threadripper 5955), sorted in descending order of model complexity. 
    For the conventional system, we used the least complex model, i.e., Fast R-CNN.
    However, utilizing our proposed module still reduces energy consumption in all models across all platforms compared to the conventional system, even if the server model is notably more complex and energy-hungry than the conventional system's server model.
    The $x$ axis is illustrated in log scale, and as presented by this figure, our proposed system consumes less than $16\%$ energy in all settings compared to the conventional system, and less than $10\%$ energy if using the same complex model as the conventional system. 

    Figure~\ref{fig: system}\textbf{d} compares the energy consumption under different ratios of background frames to FOIs ($M$).
    In the experiment, we fixed the number of FOIs and adjusted the number of background frames to tune the ratio.
    As $M$ increases, the energy consumption for all configurations rises primarily due to a corresponding rise in the number of inferences executed on the central server.
    However, compared to the conventional system's significant energy consumption increase (from the magnitude of $10^5$ to $10^6$), the increment of our system is 10 times less than that of the conventional system.
    It illustrates the scenario in which our system is more beneficial: the less frequently FOIs appear, the more energy our system can save.

\begin{figure*}[!ht]%
    \centering
    \includegraphics[width=1\textwidth]{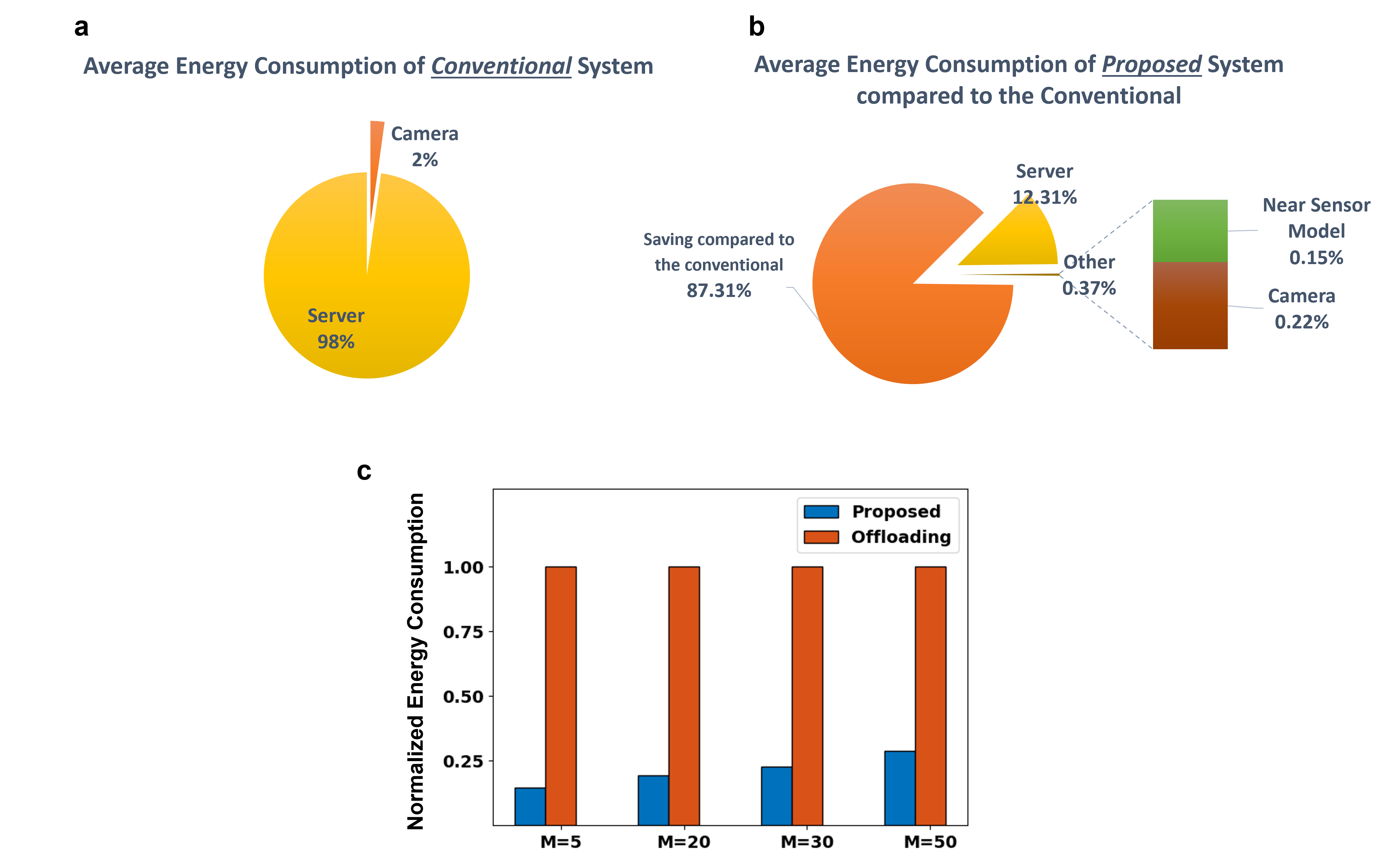}
   \caption{ \textbf{Energy consumption comparison.} \textbf{a} The average energy consumption breakdown for the conventional system. \textbf{b} The average energy consumption breakdown for the framework utilizing the proposed module compared to the conventional framework. \textbf{c} Energy consumption comparison of our proposed method and the baseline. }  
    \label{fig: comparison}
    \end{figure*}

\begin{figure}[!h]%
\centering
\includegraphics[width=1\columnwidth]{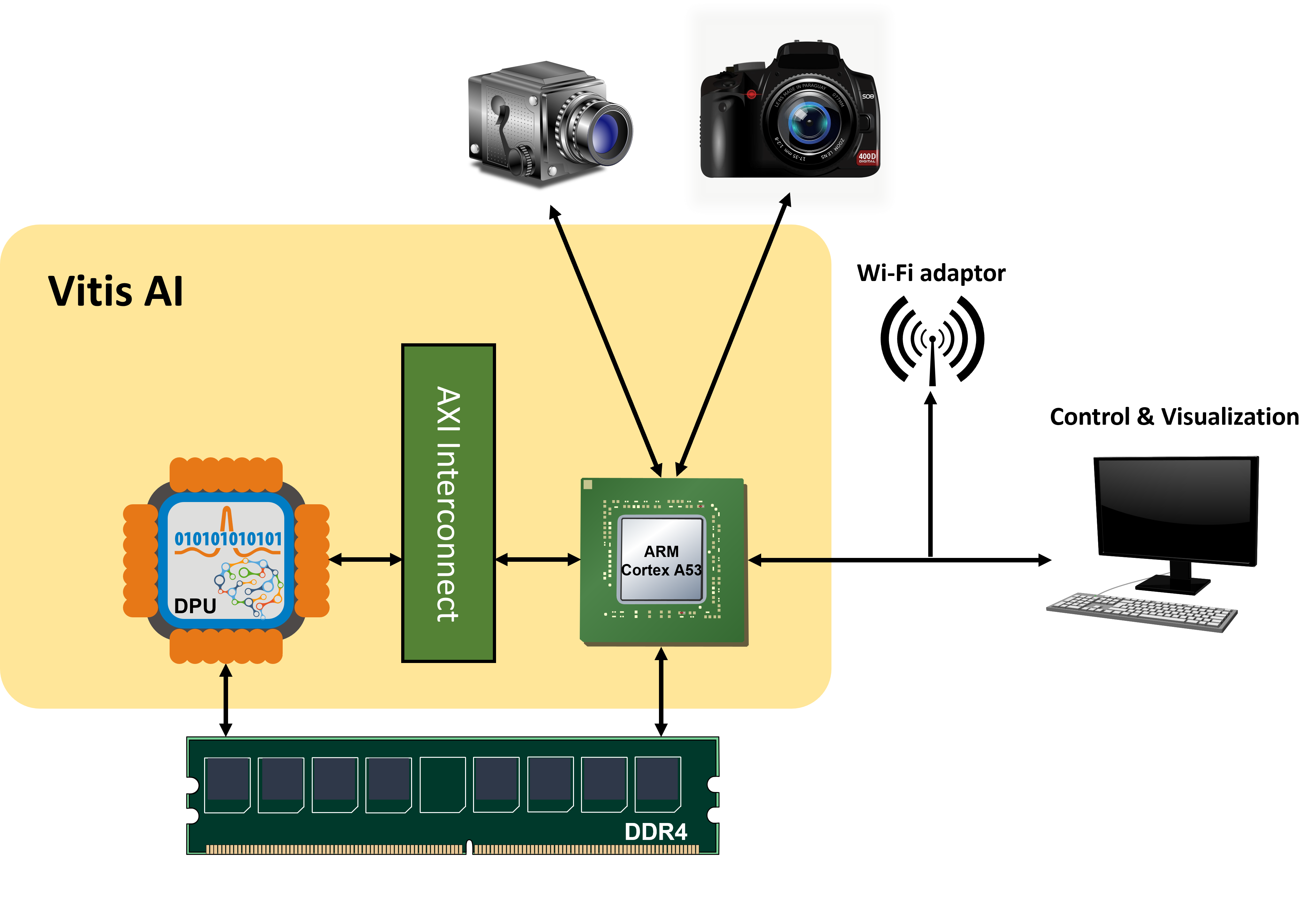}
\caption{\textbf{Hardware platform overview.}}\label{fig:hardware}
\end{figure}

    Figure 5\textbf{a} and \textbf{b} show the average energy consumption breakdown of the conventional system and our proposed system. Although our system adds a negligible energy portion to the system (from the near-sensor model), it is capable of saving $87\%$ on the energy consumption of the conventional system and keeps valuable information.
     Figure~\ref{fig: comparison}\textbf{c} depicts a comparison of systemwide energy consumption between our method and an offloading strategy designed to address energy constraints that assumes deploying the same model on the edge \cite{yu2020energy}. 
    The figure highlights the sensitivity of our method to the value of the M ratio. 
    In all scenarios, our approach significantly outperforms the baseline. 
    Specifically, it enables more obvious improvement when M is small.
    Importantly, our method is adaptable to complement existing approaches, combining their advantages effectively.

\section{Discussion}
With the rise in IoT devices and data generated by them, proposing methods to analyze this huge data with minimal energy consumption is crucial.
In this article, we introduce a novel module for intelligent sensing that addresses some of the challenges associated with analyzing large-scale sensor data using complex machine learning models. 
Our intelligent module is designed based on the observation that in many IoT applications, only a small proportion of sensor data conveys information of interest.
Clearly, it is inefficient to transmit and analyze all frames captured by the sensors, as it results in unnecessary resource consumption.
Therefore, our module intelligently selects the data generated by the sensors, and only transmits and analyzes the data with useful information.
It employs a near-sensor lightweight model to detect information of interest, a switch to control the data transmission rate, and a complex model located in the server to implement more sophisticated inference.
Specifically, the near-sensor model and switch are designed to reduce the energy and storage requirements of the entire framework, with a focus on decreasing the transmission frequency when no useful information is detected.

We evaluated the performance of our module using a visual monitoring scenario, where we identify frames of interest (FOI) as the useful information that should be transmitted to the complex model and background frames as the irrelevant information to be discarded.
On the lightweight model side, to achieve high efficiency, we utilized the YOLO family of models and investigated the performance of various model sizes.
Additionally, we customized the model architecture and the loss function to fit our scenario and implemented quantization to further compress the model.
Our customized model was able to maintain a level of performance that is comparable to the original model, even when we reduced the number of parameters to just $6.2\%$ of the original model and used int4 quantization for precision. 
We also proposed two schemes to improve the overall performance of our module and mitigate the potential misdetections of the near-sensor model.
First, instead of cutting off the data transmission completely when no FoI is detected, we considered a more conservative approach and defined a non-zero minimum transmission frequency that ensures regular transmission to the server at a low frequency even if no FOI is detected. 
Second, we introduced a lazy sensor deactivation scheme leveraging the temporal correlation between the adjacent frames.
Rather than reducing the transmission frequency immediately after detecting a background frame, our module employs a switch that maintains a high transmission frequency until a threshold number of consecutive background frames is reached. 
This threshold is automatically adjusted based on the previous detection results to strike a balance between resource consumption and accuracy.
Our experiments demonstrated the efficacy of the frameworks utilizing our proposed module on various platforms and with complex models on the server. 
Although our proposed module introduces extra energy consumption from the near-sensor model, it still can significantly decrease the total energy consumption and storage usage to less than 10\% of the conventional system without our intelligent module, while maintaining over 95\% of useful information.
As the ratio of background frames to FOIs increases, our proposed module shows more strength. 
This indicates the versatility and generalizability of our module, making it applicable to a wide range of IoT applications in which useful information happens rarely.

In summary, our article introduces a new paradigm for intelligent sensing systems by rethinking and designing the data transmission policy. 
This intelligent module has broad applicability and significant impacts on various IoT applications, including but not limited to security surveillance, environmental monitoring, and criminal alarms.

\section*{Data availability statement}

The dataset Microsoft COCO object detection for this study can be found in \cite{lin2014microsoft}. The raw data supporting the conclusion of this article will be made available by the authors, without undue reservation.

\appendices
\section{Supplementary Material}
\label{sec:sample:appendix}
\subsection*{Video Demo} 
Our research includes a video demonstration showcasing the results. In the demo, our model detects the animals appearing in the frames. The video can be accessed at the following link: 

\url{https://drive.google.com/file/d/1-IpRLfd8Ym38p8APCJgxNq5igiK5ARa5/view?usp=sharing}

\begin{algorithm}
\caption{Intelligent data transmission}\label{algoA1}
\begin{algorithmic}[1]
\Require YOLO prediction($y$), Lazy sensor deactivation count($N$), Camera refresh rate($f_r$), Minimum transmission frequency($f_{min}$), $C_1,C_2,C_3=0,0,0$
\Ensure Transmission decision($D$)
\If{$y == 1$}
        \State $C_1, C_2,C_3=0,0,0$
        \State \Return $D=1$
\Else
    \If{$C_3==0$}
        \State $C_1 = C_1 + 1$ \label{code:line6}
        \If{$C_1 \leq \max(1, \frac{N}{2^{C_2}})$}
            \State \Return $D=1$
        \Else
            \State $C_1,C_2,C_3=0,C_2+1,C_3+1$
            \State \Return $D=0$
        \EndIf \label{code:line12}
    \Else\label{code:min transmission}
        \State $C_3 = C_3 + 1$ \label{code:line14}
        \If{$C_3 == f_r/f_{min}$}
            \State $C_1,C_3=C_1+1, 0$
            \State \Return $D=1$
        \Else
            \State \Return $D=0$
        \EndIf \label{code:line19}
    \EndIf
\EndIf  
\end{algorithmic}
\end{algorithm}

\section*{Acknowledgment}

This work was supported in part by National Science Foundation \#2127780 and \#2312517, Office of Naval Research, grants \#N00014-21-1-2225 and \#N00014-22-1-2067, the Air Force Office of Scientific Research under award \#FA9550-22-1-0253, and generous gifts from Xilinx and Cisco. 
We express our gratitude to Andrew Ding for the discussion and meticulous proofreading to improve the earlier version of the manuscript.

\section*{Competing interests}
The author(s) declare no competing interests.

\ifCLASSOPTIONcaptionsoff
  \newpage
\fi
\bibliographystyle{elsarticle-num} 
\bibliography{bibtex/bib/IEEEexample}
 
\end{document}